\documentclass{article}
\usepackage{spconf,amsmath,graphicx}
\usepackage{multirow}
\usepackage{subcaption}
\usepackage{array}
\usepackage{cite}
\usepackage{amssymb}
\usepackage{mathrsfs}
\usepackage{blindtext}
\usepackage{algpseudocode}
\usepackage{algorithm}
\usepackage{multirow}
\usepackage{bm}
\usepackage{booktabs}
\usepackage{url}
\usepackage[super]{nth}

\title{HaikuS2S: A Cascaded System For Responding In Verse}

\name{
\begin{tabular}{c}
Devangi Sharma$^1$\sthanks{Equal first-authorship contribution}, Sophia Judicke$^{1*}$, Glenda Tan$^1$\sthanks{Equal second-authorship contribution}, Conrad Schaumburg$^{2\dagger}$\sthanks{Work completed at Carnegie Mellon University}, Shinji Watanabe${^1}$
\end{tabular}
}
\address{$^1$ Carnegie Mellon University \quad $^2$ Unaffiliated \\
\small{\texttt{\{sjudicke, devangis, glendat, swatanab\}@andrew.cmu.edu}, \quad \texttt{conradschaumburg@gmail.com}}
}

\usepackage{spconf}
\usepackage{tikz} 

\newcommand\copyrighttext{%
  \footnotesize \textcopyright 2026 IEEE. Personal use of this material is permitted. Permission from IEEE must be obtained for all other uses, in any current or future media, including reprinting/republishing this material for advertising or promotional purposes, creating new collective works, for resale or redistribution to servers or lists, or reuse of any copyrighted component of this work in other works.}

\newcommand\copyrightnotice{%
\begin{tikzpicture}[remember picture,overlay]
\node[anchor=south,yshift=10pt] at (current page.south) 
  {\fbox{\parbox{\dimexpr\textwidth-\fboxsep-\fboxrule\relax}{\copyrighttext}}};
\end{tikzpicture}%
}

\begin{document}
\ninept
\maketitle
\copyrightnotice
\begin{abstract}
Expressive speech synthesis has advanced through prosody modeling, yet generating structured poetic speech, such as haiku, remains challenging. Prior work on prosody transfer improves expressiveness, and fine‑tuned poetry TTS (text-to-speech) systems capture verse intonation. However, these methods do not model haiku’s 5‑7‑5 syllable structure or line-ending pauses. We present a cascaded system, HaikuS2S, combining ASR (automatic speech recognition), LLM (large language model)-generated haiku, and TTS fine-tuning on both prose and custom haiku datasets. Our evaluation focuses on emotion similarity, speech quality, and prosody alignment. In our experiments, we see that our prosody and tonal alignment improve significantly with our fine-tuned systems, particularly the one trained on both general poetry and haiku. We also see that we maintain similar emotion similarity scores across all systems.
\end{abstract}
\section{Introduction}
\label{sec:intro}
Neural TTS and prosody modeling have improved expressive speech synthesis, but generating structured poetic speech like haiku is still difficult. Standard TTS often fails to reproduce poetic rhythm, line breaks, or emotional nuances. Prosody transfer methods enable expressive speech via learned embeddings \cite{skerry-ryan18a}, and poetry TTS systems can capture verse intonation \cite{koch2022poetictts}. Emotion-aware models such as emotion2vec further enhance expressiveness \cite{ma2023emotion2vec}. Yet, existing approaches ignore haiku-specific constraints like the 5‑7‑5 syllable structure and line-ending pauses. We address this with HaikuS2S, a cascaded system that integrates ASR, LLM-based haiku generation, and TTS fine-tuning on prose and custom haiku datasets, aiming for accurate prosody, rhythm, and expressive naturalness.

The motivation for focusing specifically on haiku generation stems from the inherent structure of haiku, which provides clear, enforceable guidelines for TTS systems. Unlike free-verse poetry, where rhythm and pauses are more abstract and subjective, haiku offers a concrete framework that can be directly modeled and learned. This makes haiku an ideal test case for exploring how well TTS can adapt to highly structured poetry. Moreover, a cascaded system gave us more fine-grained control to enforce these constraints versus an end-to-end system. The 5-7-5 format of a haiku provides an easy pattern for a TTS to identify; therefore, this work explores how varying domain specificity in training data impacts TTS performance on haiku pronunciation. This study investigates whether a generic poetry dataset is sufficient to capture haiku recitation. Our work builds on the intersection of expressive prosody control, cascaded systems, and domain-specific fine-tuning, offering a new approach to creating expressive and accurate synthetic speech for structured poetic forms. Our contributions also include making our final system model and training and testing data publicly available via HuggingFace\footnote{\url{https://huggingface.co/HaikuS2S}}.

For HaikuS2S to succeed in haiku performance, we want it to correctly capture the prosody of a given haiku. This includes pausing on line breaks and correctly pausing, stopping, or changing tone on different punctuation marks. We also ensure our LLM outputs a valid 5-7-5 haiku.

\section{Related studies}
\label{sec: related works}
Research on expressive text‑to‑speech and prosody modeling has grown rapidly in recent years.

\textbf{Prosody Transfer and Control:} Prosody transfer techniques learn latent prosody embeddings that enable TTS models to replicate prosodic patterns from reference audio \cite{skerry-ryan18a}. Approaches have been proposed for fine‑grained prosody control by modeling prosodic features such as duration and fundamental frequency for expressive TTS \cite{pamisetty2022prosodytts}, enabling control of prosody at fine levels.  

\textbf{Emotion and Prosody Representation:} Modeling emotional content in speech has become crucial for expressive TTS. Self‑ supervised speech representation models like emotion2vec capture emotion across multiple languages and tasks \cite{ma2023emotion2vec}. Multi‑scale emotion modeling and control frameworks further enhance expressive synthesis by representing both global and local prosodic variations \cite{lei2022msemo}, while disentanglement approaches separate prosody and timbre for better cross‑speaker emotion transfer \cite{zhang2022iemotts}.

\textbf{Poetry Synthesis and Prosodic Structure:} PoeticTTS demonstrates that fine‑tuning on verse data yields improved poetic intonation in TTS outputs \cite{koch2022poetictts}. SPARSAR presents a computational poetry analysis pipeline which combines syntactic, semantic, and prosodic information for expressive poetry reading \cite{delmonte2014sparsar}. Semantics for expressive TTS explores how linguistic semantics contribute to naturalness and expressiveness in synthesized speech \cite{wang2015semantics}. However, these methods do not explicitly enforce strict constraints like haiku’s 5‑7‑5 syllable structure and poetic pauses.

\textbf{Structured Data and Low‑Resource TTS:} Low‑resource expressive synthesis research highlights the challenges of modeling prosody and emotion with limited training data \cite{lowresource2020tts}. End‑to‑end models focusing on expressive prosody across multi‑sentence contexts show that longer context and richer text features improve prosodic coherence \cite{makarov2022simpletts}.  

Together, these studies highlight advances in expressive TTS, prosody modeling, and poetic synthesis. Our work builds on this literature by combining expressive prosody control, emotion representation, and domain‑specific fine‑tuning on haiku data for structured poetic speech synthesis. Restricting to a cascaded system and focusing on haiku-specific constraints made the problem easier to tackle, as notions of prosody and success were less abstract and had more explicit rules.

\section{System Design and Method}
\label{sec: problem formulation}
HaikuS2S takes the form of a single-speaker cascaded system (see Fig. 1). We use ESPnet \cite{espnet} \cite{espnetSDS} framework with \verb|espnet/owsm_| \verb|v4_base_102M| as our ASR model \cite{owsm-v4}, \verb|microsoft/| \verb|Phi-3| \verb|mini-4k-instruct| \cite{abdin2024phi3} as our LLM, and \verb|kan-bayashi/| \verb|vctk_multi_spk_vits| as our TTS model \cite{hayashi2020espnet}. We believe that a pretrained-only model, such as the mentioned multispeaker VITS model, provides an accurate baseline to see how a cascaded system performs at haiku recitations, as the model is trained only on novel-based and conversational texts. In our LLM prompt, we include directions to respond to the user in haiku format: 
\begin{quote}
\itshape 
``You are a haiku poet. Reply with exactly one haiku in English: three lines with a 5-7-5 syllable pattern (five syllables, then seven, then five). No title, no preamble, no numbering—only the three lines. The haiku should reflect or respond to the user's words.''
\end{quote}

We have two experimental stages of HaikuS2S, which use the same ASR model, LLM, and LLM prompt. We follow two different TTS fine-tuning stages:
\begin{enumerate}
    \item \textbf{Poetry fine-tuning} on the \verb|mythicinfinity/libritts| \verb|_r| \cite{Koizumi2023-hs} dataset of standard poetry readings, improving general prosody, intonation, and naturalness for spoken poetry.
    \item \textbf{Haiku fine-tuning} on our custom haiku dataset, enforcing haiku-specific features such as 5-7-5 syllable structure and line-ending pauses.
\end{enumerate}
Our proposed methods solve our task of responding to user queries in spoken haiku format by aiding in the prosody of haiku responses. Stage 2 is unique and distinct from prior results because, to our knowledge, nobody has performed hyperspecific fine-tuning on haiku data for the purpose of haiku reading correctness and tonality, and prior work has not combined strict poetic form enforcement with prosody-aware TTS fine-tuning in a cascaded generation pipeline. This two-stage fine-tuning will introduce prosodic control in stage 1, allowing for better convergence at stage 2 despite a very small stage 2 training set.

\begin{figure}[t]
    \centering
    \includegraphics[width=0.5\textwidth]{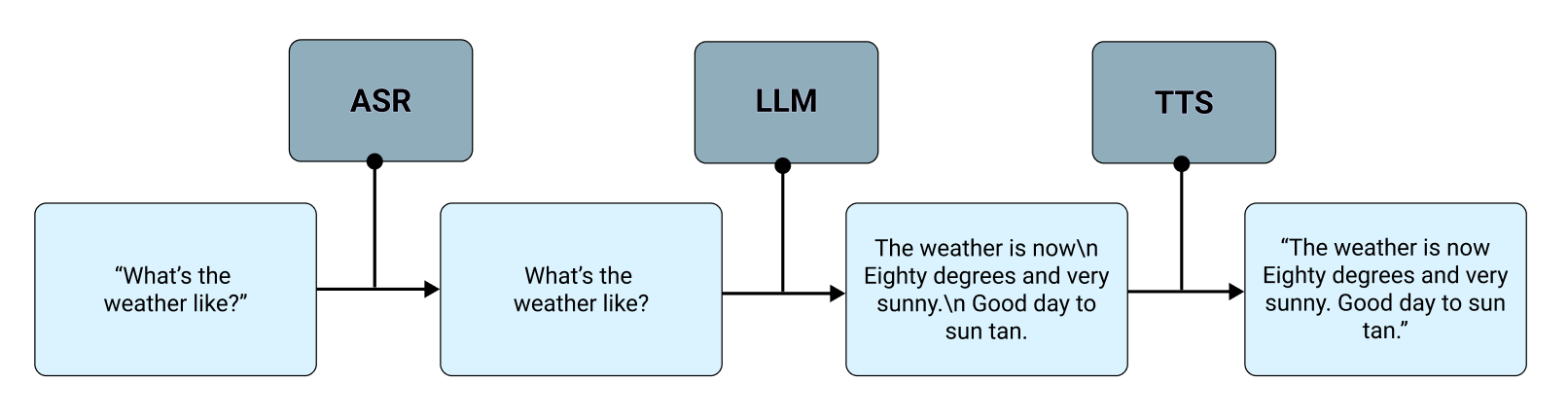}
    \caption{Block diagram showing our cascaded haiku spoken dialog system deployed in ESPnet-SDS. Displays example user query and example haiku response.}
    \label{fig:fig1}
\end{figure}

HaikuS2S ensures correct 5-7-5 haiku formatting via G2P (grapheme-to-phoneme) validation, which counts the number of vowels in each line, after a haiku is generated. This occurs between the LLM and TTS steps, where we prompt the LLM to regenerate the haiku until it satisfies the 5-7-5 format. We also introduce a special token \verb|[\n]| in stages 1 and 2 (but not the baseline) to indicate new lines, which we hypothesize will implicitly enable better prosodic control. This token should allow the model to learn to pause for the appropriate length following each line, mimicking natural recitation. The token was left as a unique token in the TTS dictionary and served as an indication of line pauses in the haiku audio files, allowing for longer pauses compared to the standard newline token. This system is deployed with Gradio on a T4 GPU \cite{Abid2019GradioHS}.

\section{Experiments}

\subsection{Datasets}
\textbf{Training Data:} Our first round of training used 827 poetry samples from the HuggingFace dataset \verb|mythicinfinity/libritts| \verb|_r|, which we selected via dataset curation. For our second round of training, we created a dataset of 500 haiku by having an LLM generate 500 haiku of varying topics. We then had two members of our team each record 250 distinct haiku in quiet settings using cell phones at 16kHz.

\textbf{Test Data:} We created a dataset of 498 haiku by having an LLM generate 500 haiku about varying topics (of which 2 were later removed for not following a 5-7-5 haiku structure). We then had two members of our team each record 250 of the haiku in a quiet setting using cell phones at 16kHz. These team members were different from those who had recorded the training data.

\subsection{Metrics}
For all our objective evaluations, we used the VERSA toolkit \cite{shi2024versaversatileevaluationtoolkit}\cite{shi2025versa}. The accuracy and intelligibility metrics we used are \verb|WER| (word-error rate) and \verb|CER| (character error rate). These metrics were scored by performing ASR (model: \verb|owsm_large-v3|) on our TTS outputs.

The tone and prosody metrics we used are the following:

\verb|emo_similarity|: This metric compares the emotions of the reference speech to the emotions of the output audio within a range of $(-1,1)$,  where $1$ is more similar and $-1$ is less similar.

\verb|nisqa|: An estimate of conversational speech quality and naturalness within a range of $(1, 5)$, where $1$ is poor, and $5$ is excellent. From this metric set, we focused on \verb|mos|, \verb|noi|, and \verb|col| metrics, because these indicate the overall (1) naturalness, (2) noisiness, and (3) coloration of the output compared to the reference audio.

\verb|chroma_alignment|: This metric compares the alignment of pitch, palette, and harmonic content of input text and output audio within non-negative numbers, where $0$ is more similar and higher numbers are less similar. From this metric set, we focused on three unscaled \verb|cosine_dtw| metrics to indicate our normalization of pitch and consistency of melody in cadence for poetry readings.

\verb|Human Evaluation|: This metric indicates the human-perceived naturalness, speed, tone, cadence, rhythm, and enjoyment, and likelihood of reuse (\verb|lor|) for each system. We gathered this data asynchronously across 35 subjects on a Likert (1–5) scale. This study was IRB-approved and utilized general listeners with no specialty in haiku or poetry.

These metrics are appropriate because they compare the emotional and tonal aspects one would expect when reciting haiku, as well as the timing of the speech itself. By assessing the above characteristics of output speech, we believe that we are measuring the most important aspects of poetry recitation.

\subsection{Experimental Setups}
We implement a two-stage fine-tuning strategy for the TTS model, defined in Section 3.

\textbf{Training details:} For both stages, we train the TTS model with a batch size of 16 using the Adam optimizer. For stage 1, the TTS model was trained for 150 epochs with a learning rate of 1e-4; for stage 2, for 600 epochs and a learning rate of 5e-4. 

\textbf{Testing details:} For the baseline and both stages, we evaluate on the test set defined in Section 4.1.

\subsection{Results and Discussion}
\textbf{Speaker Intelligibility:} Stage 2 had the lowest error rate across all three systems. Stage 2 achieved the lowest error rates (Fig. 2), with a more pronounced improvement in WER than CER. We hypothesize that this is because rhythm is taken into account by \verb|owsm_wer| and \verb|owsm_cer|. Since our baseline and stage 1 systems do not identify pauses with the same level of correctness as stage 2 of HaikuS2S, it is expected that they are considered less intelligible. 
\textbf{Emotion Similarity:} All systems maintained a calm emotional tone, though Stage 2 yielded statistically significant improvements in matching the exact target emotional similarity (Fig. 3).
\begin{figure}[h]
    \centering
    \includegraphics[width=0.5\textwidth]{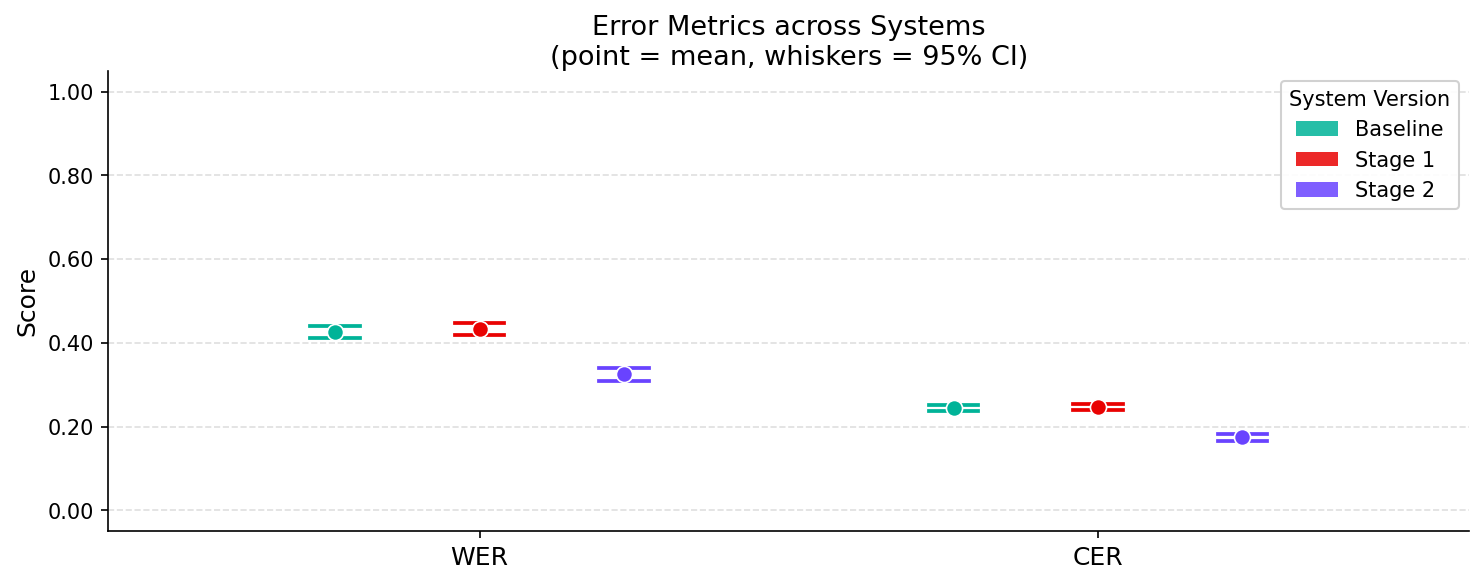}
    \caption{Plots displaying word and character error rates across HaikuS2S stages. A lower score indicates greater correctness.}
    \label{fig:fig2}
\end{figure}
\vspace{-15pt}
\begin{table}[h]
    \centering
    \begin{tabular}{c|c}
        \hline
        \textbf{Stage} & \textbf{Emotional Similarity} \\ \hline
        Baseline & 0.935 $\pm$ 0.0012 \\
        Stage 1 & 0.946 $\pm$ 0.0010 \\
        Stage 2 & 0.954 $\pm$ 0.0008 \\
        \hline
    \end{tabular}
    \caption{Emotion similarity across HaikuS2S stages. A higher score indicates more similarity with the golden standard audio.}
    \label{tab:tab1}
\end{table}
\vspace{-10pt} \\
\textbf{Naturalness and Speech Quality:} All three systems (baseline, stage 1, and stage 2) had statistically significant differences in NISQA metrics. We see that the baseline and stage 1 consistently performed better than stage 2 (Fig. 4). We hypothesize that stage 2 performs poorly compared to the baseline and stage 1 because NISQA measures conversational naturalness, while stage 2 outputs more strictly haiku-structured audio. Since NISQA penalizes long pauses or unnatural conversational cadences and noise-level changes, which could include poetry line breaks, it is somewhat expected that stage 2 performs relatively poorly.
\textbf{Chroma Alignment: } All three systems (baseline, stage 1, and stage 2) had statistically significant differences in chroma alignment. We see that stage 2 outperforms the baseline and stage 1 on two of the three metrics (Fig. 5). We also see that stage 1 performs equally or better compared to the baseline. We hypothesize that stage 1 is better than stage 2 on \verb|chroma_stft| because this metric penalizes monotonicity, and we found that stage 2 outputted slightly more monotone audio outputs. We believe that the baseline consistently performs the worst because the baseline model is not meant to read poetry or prose. In contrast, stage 1 is trained on general poetry, and stage 2 is trained on poetry and also haiku with correct vocal tone, leading to higher similarity of output audios to gold standard audios.
\begin{figure}[H]
    \centering
    \includegraphics[width=0.5\textwidth]{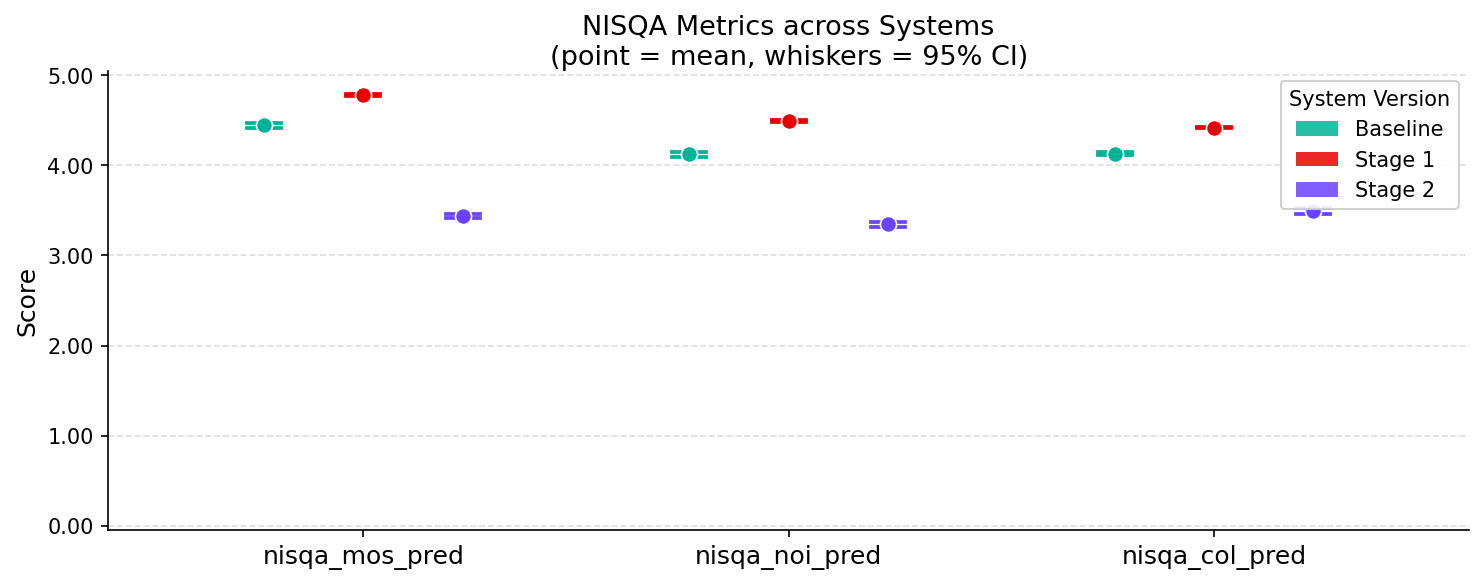}
    \caption{Plots displaying NISQA metrics across HaikuS2S stages. A higher score indicates more natural-sounding speech.}
    \label{fig:fig4}
\end{figure}
\vspace{-15pt}
\begin{figure}[H]
    \centering
    \includegraphics[width=0.5\textwidth]{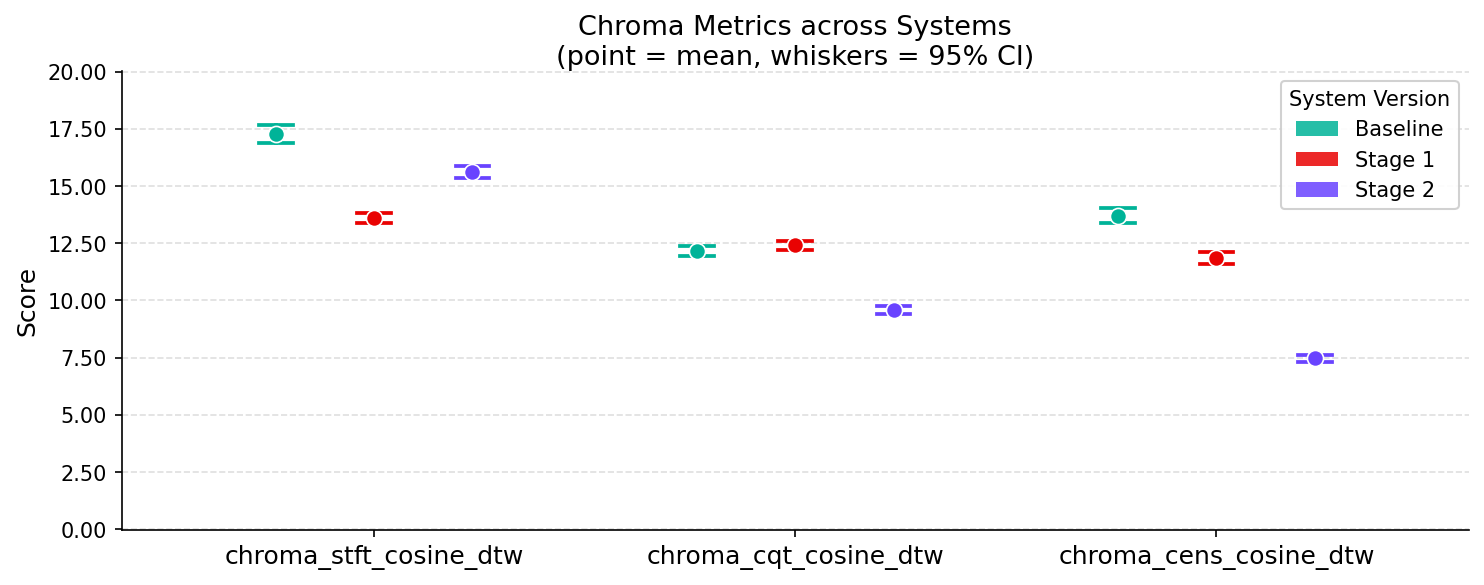}
    \caption{Plots displaying chroma metrics across HaikuS2S stages. A lower score indicates more similarity with the golden audios.}
    \label{fig:fig5}
\end{figure}
\vspace{-15pt} 
\begin{figure}[H]
    \centering
    \includegraphics[width=0.5\textwidth]{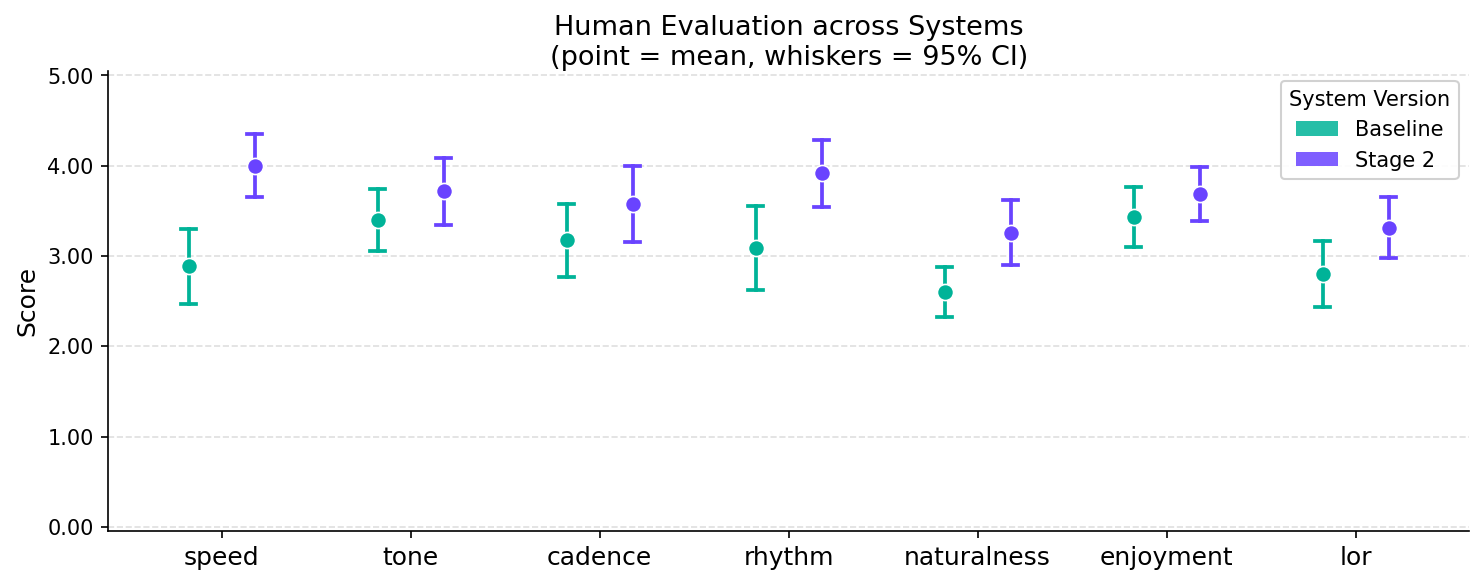}
    \caption{Plots displaying human evaluation across HaikuS2S stages. A higher score indicates higher preference of system.}
    \label{fig:fig6}
\end{figure}
\vspace{-10pt}
\textbf{Human Evaluation}: The baseline and stage 2 had higher scores on average across all evaluation metrics on a very small sample size of 35 (Fig. 6, 7). This is anticipated, as the baseline is not trained on poetry or haiku data, while stage 2 is trained on both. This leads to a higher human rating, as audio outputs are more similar to a recitation of a haiku by a human in stage 2 compared to the baseline. We argue that this result is important as this is the only metric we have that captures correct pronunciation and naturalness of poetry audios.
\textbf{Overall Comparison}: We argue that stage 2 performs the best of the three systems. Combining intelligibility, tone, alignment, and human-preferred audios, we achieve our goal of creating a system that can correctly recite haiku. This performance increase is evident in both the metrics and the output audio for a human listener.

We also argue that stage 1 performs better than the baseline. Similarly, we see a notable improvement in the prosodic abilities of stage 2, which is clear to a human listener.

\textbf{Future Work:} In the future, it would be interesting to see if we can train the TTS to mimic the 5-7-5 constraints using G2P alignment decoding. We would also be interested in exploring the abilities of HaikuS2S on other kinds of poetry, such as sonnets and free-verse, prose, and general audiobooks. Further, we wish to develop a poetry-specific evaluation metric to augment or replace NISQA for more robust evaluation.

\clearpage
\section{AI-Generated Content Disclosure}
\vspace{-5pt}
We used Gemini to generate our 1000 haiku (both training and testing data) used in the dataset \verb|1000-haiku-audio|. We also used Claude to aid in debugging fine-tuning scripts. No AI-generated content was used in the text, figures, or images included in this article.
\vspace{-10pt}

\nocite{*}
\bibliographystyle{IEEEbib}
\bibliography{refs}
\vspace{-10pt}

\newpage
\section{Appendix: Supplementary Tables}
\vspace{-16pt}
\begin{table}[h]
    \centering
    \begin{tabular}{c|c|c}
        \hline
        \textbf{Stage} & \textbf{WER} & \textbf{CER} \\ \hline
        Baseline & 0.425 $\pm$ 0.013 & 0.244 $\pm$ 0.007 \\
        Stage 1 & 0.433 $\pm$ 0.014 & 0.081 $\pm$ 0.007 \\
        Stage 2 & 0.324 $\pm$ 0.014 & 0.099 $\pm$ 0.008 \\
        \hline
    \end{tabular}
    \caption{WER and CER across HaikuS2S stages.}
    \label{tab:wer_cer}
\end{table}
\vspace{-19pt}
\begin{table}[h]
    \centering
    \begin{tabular}{c|c|c|c}
        \hline
        \textbf{Stage} & \textbf{MOS} & \textbf{NOI} & \textbf{COL} \\ \hline
        Baseline & 4.442 $\pm$ 0.024 & 4.120 $\pm$ 0.028 & 4.126 $\pm$ 0.015 \\
        Stage 1 & 4.78 $\pm$ 0.011 & 4.490 $\pm$ 0.013 & 4.416 $\pm$ 0.005\\
        Stage 2 & 3.439 $\pm$ 0.020 & 3.343 $\pm$ 0.023 & 3.488 $\pm$ 0.032\\
        \hline
    \end{tabular}
    \caption{NISQA metrics across HaikuS2S stages.}
    \label{tab:nisqa}
\end{table}
\vspace{-19pt}
\begin{table}[h]
    \centering
    \begin{tabular}{c|c|c|c}
        \hline
        \textbf{Stage} & \textbf{STFT} & \textbf{CQT} & \textbf{CENS} \\ \hline
        Baseline & 17.276 $\pm$ 0.410 & 12.169 $\pm$ 0.219 & 13.710 $\pm$ 0.326 \\
        Stage 1 & 12.601 $\pm$ 0.232 & 12.403 $\pm$ 0.175 & 11.863 $\pm$ 0.266\\
        Stage 2 & 15.623 $\pm$ 0.256 & 9.599 $\pm$ 0.176 & 7.466 $\pm$ 0.135\\
        \hline
    \end{tabular}
    \caption{Chroma metrics across HaikuS2S stages.}
    \label{tab:chroma}
\end{table}
\vspace{-19pt}
\begin{table}[H]
    \centering
    \label{tab:human_evaluation}
    \resizebox{\columnwidth}{!}{%
    \begin{tabular}{c|c|c|c}
        \hline
        \textbf{Stage} & \textbf{Speed} & \textbf{Tone} & \textbf{Cadence} \\ \hline
        Baseline & $2.886 \pm 0.415$ & $3.400 \pm 0.342$ & $3.171 \pm 0.405$ \\
        Stage 2  & $4.000 \pm 0.350$ & $3.714 \pm 0.373$ & $3.571 \pm 0.419$ \\ \hline
        \hline
        \textbf{Rhythm} & \textbf{Naturalness} & \textbf{Enjoyment} & \textbf{LOR} \\ \hline
        $3.086 \pm 0.464$ & $2.600 \pm 0.280$ & $3.429 \pm 0.334$ & $2.800 \pm 0.366$ \\
        $3.914 \pm 0.371$  & $3.257 \pm 0.362$ & $3.686 \pm 0.298$ & $3.314 \pm 0.338$ \\ \hline
    \end{tabular}%
    }
    \caption{Human evaluation across HaikuS2S stages.}
\end{table}

\end{document}